\documentclass[runningheads]{llncs}
\usepackage{graphicx}
\usepackage[utf8]{inputenc} 
\usepackage[T1]{fontenc}    
\usepackage{hyperref}       
\usepackage{url}            
\usepackage{booktabs}       
\usepackage{amsfonts}       
\usepackage{nicefrac}       
\usepackage{microtype}      
\usepackage{lipsum}
\usepackage{tikz}
\usepackage{pgfplots}
\usetikzlibrary{pgfplots.groupplots}
\pgfplotsset{width=7cm, compat=1.14}
\usepgfplotslibrary{external}
\usepackage{graphicx, upgreek, tikz, standalone}
\usetikzlibrary{arrows, decorations, graphs, decorations.pathmorphing, decorations.pathreplacing, decorations.shapes, backgrounds, positioning, fit, petri, bending, intersections, automata, shapes.misc, calc, arrows.meta}

\usepackage[ruled,vlined,algo2e]{algorithm2e}
\usepackage{algorithm, algorithmic}

\begin{document}
\title{Increasing the Inference and Learning Speed of Tsetlin Machines with Clause Indexing}
\titlerunning{Tsetlin Machines with Clause Indexing}
%

\author{
Saeed Rahimi Gorji\inst{1} \orcidID{0000-0002-2699-9903} \and
Ole-Christoffer Granmo\inst{1}
\orcidID{0000-0002-7287-030X}
\and
Sondre Glimsdal\inst{1} \and
Jonathan Edwards\inst{2}
\and Morten Goodwin\inst{1} \orcidID{0000-0002-7287-030X}
}
\authorrunning{S. R. Gorji et al.}

\institute{Centre for Artificial Intelligence Research, University of Agder, Grimstad, Norway
\and
Temporal Computing, Newcastle, United Kingdom\\
\email{
saeed.r.gorji@uia.no, ole.granmo@uia.no, sondre.glimsdal@uia.no, jonny@temporalcomputing.com, morten.goodwin@uia.no
}\\
}
\maketitle
\begin{abstract}
The Tsetlin Machine (TM) is a machine learning algorithm  founded on the classical Tsetlin Automaton (TA) and game theory. It further leverages frequent pattern mining and resource allocation principles to extract common patterns in the data, rather than relying on minimizing output error, which is prone to overfitting. Unlike the intertwined nature of pattern representation in neural networks, a TM decomposes problems into self-contained patterns, represented as conjunctive clauses. The clause outputs, in turn, are combined into a classification decision through summation and thresholding, akin to a logistic regression function, however, with binary weights and a unit step output function. In this paper, we exploit this hierarchical structure by introducing a novel algorithm that avoids evaluating the clauses exhaustively. Instead we use a simple look-up table that indexes the clauses on the features that falsify them. In this manner, we can quickly evaluate a large number of clauses through falsification, simply by iterating through the features and using the look-up table to eliminate those clauses that are falsified. The look-up table is further structured so that it facilitates constant time updating, thus supporting use also during learning. We report up to 15 times faster classification and three times faster learning on MNIST and Fashion-MNIST image classification, and IMDb sentiment analysis.

\end{abstract}

\section{Introduction}

The Tsetlin Machine (TM) is a recent machine learning approach that is based on the Tsetlin Automaton (TA) by M.~L. Tsetlin \cite{Tsetlin1961}, one of the pioneering solutions to the well-known multi-armed bandit problem \cite{Robbins1952,Gittins1979} and the first Learning Automaton \cite{Narendra1989}. It further leverages frequent pattern mining \cite{Haugland2014} and resource allocation \cite{Granmo2007d} to capture frequent patterns in the data, using a limited number of conjunctive clauses in propositional logic. The clause outputs, in turn, are combined into a classification decision through summation and thresholding, akin to a logistic regression function, however, with binary weights and a unit step output function \cite{berge2019}. Being based on disjunctive normal form (DNF), like Karnaugh maps, the TM can map an exponential number of input feature value combinations to an appropriate output \cite{phoulady2020weighted}.

\subsubsection{Recent Progress on TMs.} Recent research reports several distinct TM properties. The clauses that a TM produces have an interpretable form (e.g., \textbf{if} X \textbf{satisfies} condition A \textbf{and not} condition B \textbf{then} Y = 1), similar to the branches in a decision tree \cite{berge2019}. For small-scale pattern recognition problems, up to three orders of magnitude lower energy consumption and inference time has been reported, compared to neural networks alike \cite{wheeldon2020hardware}. The TM structure captures regression problems, comparing favourably with Regression Trees, Random Forests and Support Vector Regression \cite{abeyrathna2019nonlinear}. Like neural networks, the TM can be used in convolution, providing competitive memory usage, computation speed, and accuracy on MNIST, F-MNIST and K-MNIST, in comparison with simple 4-layer CNNs, K-Nereast Neighbors, SVMs, Random Forests, Gradient Boosting,  BinaryConnect, Logistic Circuits, and ResNet  \cite{granmo2019convtsetlin}. By introducing clause weights that allow one clause to represent multiple, it has been demonstrated that the number of clauses can be reduced up to $50\times$, without loss of accuracy, leading to more compact clause sets \cite{phoulady2020weighted}. Finally, hyper-parameter search can be simplified with multi-granular clauses, eliminating the pattern specificity parameter \cite{gorji2019multigranular}.

\subsubsection{Paper Contributions and Organization.} The expression power of a TM is governed by the number of clauses employed. However, adding clauses to increase accuracy comes at the cost of linearly increasing computation time. In this paper, we address this problem by exploiting the hierarchical structure of the TM. We first cover the basics of TMs in Sect. \ref{sec:introduction}. Then, in Sect. \ref{sec:indexing}, we introduce a novel indexing scheme that speeds up clause evaluation. The scheme is based on a simple look-up tables that indexes the clauses on the features that falsify them. We can then quickly evaluate a large number of clauses through falsification, simply by iterating through the features and using the look-up table to eliminate those clauses that are falsified. The look-up table is further structured so that it facilitates constant time updating, thus supporting indexing also during TM learning. In Sect. \ref{sec:empirical_results}, we report up to 15 times faster classification and three times faster learning on three different datasets: IMDb sentiment analysis, and MNIST and Fashion-MNIST image classification. We conclude the paper in Sect. \ref{sec:conclusion}. 

\section{Tsetlin Machines}\label{sec:introduction}
The Tsetlin Machine (TM)\cite{granmo2018tsetlin} is a novel machine learning approach that is particularly suited to low power and explainable applications \cite{wheeldon2020hardware,berge2019}. 
The structure of a TM is similar to the majority of learning classification algorithms, with forward pass inference and backward pass parameter adjustment (Fig. \ref{fig1}). A key property of the approach is that the whole system is Boolean, requiring an encoder and decoder for any other form of data. Input data is defined as a set of feature vectors:
$D= \{\mathbf{x}_1, \mathbf{x}_2,\ldots \}$,
and the output $Y$ is formulated as a function parameterized by the actions $A$ of a team of TAs: $Y = f(x;A)$.

\subsubsection{TA Team.}  The function $f$ is parameterized by a team of TAs, each deciding the value of a binary parameter $a_k \in A$ (labelling the full collection of parameters $A$). Each TA has an integer state $t_k \in \{1, \ldots, 2N\}$, with $2N$ being the size of the state space. The action $a_k$ of a TA is decided by its state: $a_k = 1 \: \mathbf{if} \: t_k > N \: \mathbf{else} \: 0.$

\subsubsection{Classification Function.} The function $f$ is formulated as an ensemble of $n$ conjunctive clauses, each clause controlling the inclusion of a feature and (separately) its negation. A clause $C_j$ is defined as:
\begin{equation}
C_j(\mathbf{x}) = \bigwedge\limits^{o}_{k=1}(x_k \lor \lnot a_k^j) \land (\neg x_k \lor \lnot a_{o+k}^j).
\end{equation}
Here, $x_k$ is the $k$th feature of input vector $\mathbf{x} \in D_o$, $o$ is the total number of features and $a_k^j$ is the action of the TA that has been assigned to literal $k$ for clause $j$.
A variety of ensembling methods can be used -- most commonly a unit step function, with half the clauses aimed at inhibition:   
\begin{equation}\label{ex}
\hat{y} = 1   \quad \mathbf{if}  \quad \left(\sum_{j=1}^{n/2} C_j^+(\mathbf{x}) -\sum_{j=1}^{n/2} C_j^-(\mathbf{x})\right) \ge 0  \quad  \mathbf{else}  \quad 0. \end{equation}
When the problem is multi-classed (one-hot encoded [15]) thresholding is removed and the maximum vote is used to estimate the class:
\begin{equation}
\hat{y} = \mathbf{argmax}_{i}\left(\sum_{j=1}^{n/2} C_{j}^{i+}(\mathbf{x}) - \sum_{j=1}^{n/2} C_{j}^{i-}(\mathbf{x})\right),
\end{equation}
with $i$ referring to the class that the clauses belong to.

\subsubsection{Learning.} Learning controls the inclusion of features or their negation. A TA $k$ is trained using rewards and penalties. If $t_k > N$ (action $1$), a reward increases the state, $t_k \mathrel{+}= 1$, up to $2N$, whilst a penalty decreases the state, $t_k \mathrel{-}= 1$. If $t_k <= N$ (action 0), a reward decreases the state, $t_k \mathrel{-}= 1$, down to $1$, whilst a penalty increases the state, $t_k \mathrel{+}= 1$. The learning rules are described in detail in \cite{granmo2018tsetlin}, the basic premise being that individual TA can be controlled by examining their effect on the clause output relative to the desired output, the clause output, the polarity (an inhibitory or excitory clause), and the action. For example, if (i) a feature is $1$, (ii) the TA state governing the inclusion of the feature is resolved to action $1$, and (iii) the overall clause outputs $1$ when a $1$ output is desired, then the TA is rewarded, because it is trying to do the right action.

\begin{figure}
\centering
\includegraphics[width=12cm]{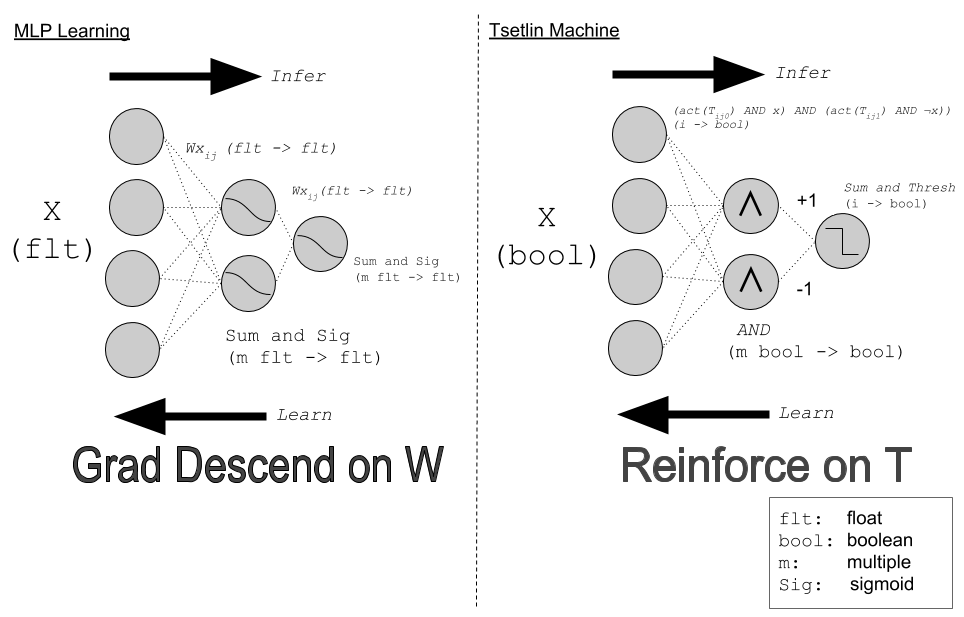}
\caption{The backwards pass requires no gradient methods (just manipulation of all TA states ($t_k$)).
Due to the negation of features, clauses can singularly encode non-linearly separable boundaries.
There are no sigmoid style functions, all functions are compatible with low level hardware implementations.}\label{fig1}
\end{figure}

The learning approach has the following added randomness:
\begin{itemize}
\item An annealing style cooling parameter $T$ regulates how many clauses are activated for updating. In general, increasing $T$ together with the number of clauses leads to an increase in accuracy.
\item A sensitivity $s$ which controls the ratio of reward to penalty, based on a split of the total probability, with values $1/s$ and $1-1/s$. This parameter governs the fine-grainedness of the clauses, with a higher $s$ producing clauses with more features included.
\end{itemize}
When comparing the TMs with a classical neural network \cite{rum} (Fig. \ref{fig1}), noteworthy points include: Boolean/integer rather than floating point internal representation; logic rather than linear algebra; and binary inputs/outputs.

\section{Fast Clause Falsification Through Indexing}\label{sec:indexing}

In this section we describe the details of our clause indexing scheme, including index based inference and learning.

\subsubsection{Overall Clause Indexing Strategy.} Recall that each TM clause is a conjunction of literals. The way that the TM determines whether to include a literal in a clause is by considering the action of the corresponding TA. Thus, to successfully evaluate a clause, the TM must scan through all the actions of the team of TAs responsible for the clause, and perform the necessary logical operations to calculate its truth value. However, being a conjunction of literals, it is actually sufficient to merely identify a single literal that is false to determine that the clause itself is false. Conversely, if no false literals are present, we can conclude that the clause is true. Our approach is thus to maintain a list for each literal that contains the clauses that include that literal, and to use these lists to speed up inference and learning.


\begin{figure}[!!t]
\begin{center}
\includegraphics[width=12cm]{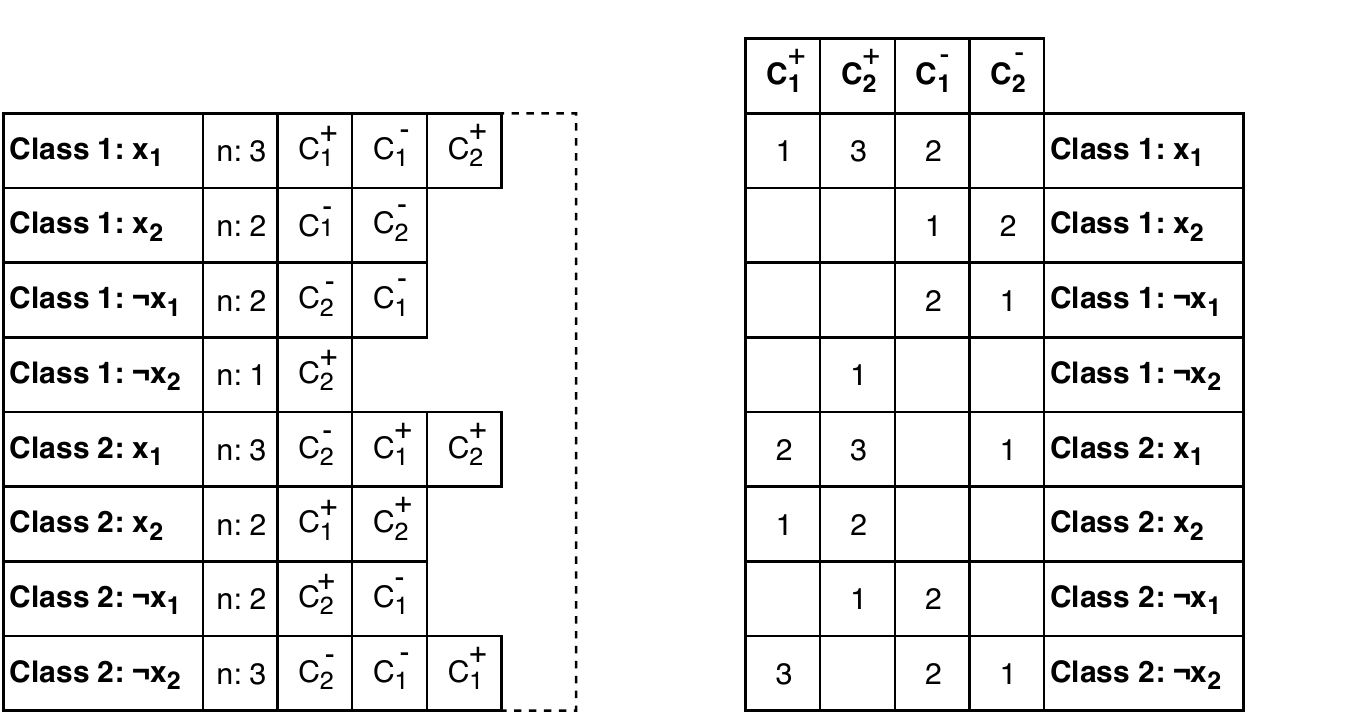}
\end{center}
\caption{Indexing structure.}
\label{fig-index}
\end{figure}

\subsubsection{Index Based Inference.} We first consider the clause lists themselves (Fig. \ref{fig-index}, left), before we propose a position matrix that allows constant time updating of the clause lists (Fig. \ref{fig-index}, right). For each class-literal pair $(i, k)$, we maintain an \emph{inclusion list} $L^i_k$ that contains all the clauses in class $i$ that include the literal $l_k$. Let the cardinality of $L^i_k$ be $|L^i_k| = n^i_k$. Evaluating the complete set of class $i$ clauses then proceeds as follows. Given a feature vector $\mathbf{x}=(x_1, x_2, \ldots, x_o)$, we define the index set $L_F$ for the false literals as $L_F = \{k|l_k = 0\}$. Then $\mathcal{C}_F^i = \bigcup_{k \in L_F} L^i_k$ is the clauses of class $i$ that have been falsified by $\mathbf{x}$. We partition $\mathcal{C}^i_F$ into clauses with positive polarity and clauses with negative polarity: $\mathcal{C}^i_F = \mathcal{C}^{i+}_F \cup \mathcal{C}^{i-}_F$. The predicted class can then simply be determined based on the cardinality of the sets of falsified clauses:
\begin{equation}
\hat{y} = \mathbf{argmax}_{i}\left(|\mathcal{C}^{i-}_F| - |\mathcal{C}^{i+}_F|\right),
\end{equation}
Alongside the inclusion list, an $m \times n \times 2o$ matrix $\mathcal{M}$, with entries $M^{ij}_k$, keeps track of the position of each clause $j$ in the inclusion list for literal $k$ (Fig. \ref{fig-index}, right). With the help of $\mathcal{M}$, searching and removing a clause from a list can take place in constant time.

\subsubsection{Index Construction and Maintenance.} Constructing the inclusion lists and the accompanying position matrix is rather straightforward since one can initialize all of the TAs to select the exclude action. Then the inclusion lists are empty, and, accordingly, the position matrix is empty as well.

Insertion can be done in constant time. When the TA responsible for e.g. literal $l_k$ in clause $C_j^i$ changes its action from exclude to include, the corresponding inclusion list $L^i_k$ must be updated accordingly. Adding a clause to the list is fast because it can be appended to the end of the list using the size of the list $n^i_k$: 
\vspace{2mm}
\begin{algorithmic}
\centering
\STATE $n^i_k \gets n^i_k+1$
\vspace{2mm}
\STATE $L^i_k[n^i_k] \gets C^i_j$
\vspace{2mm}
\STATE $M^{ij}_k \gets n^i_k$.
\end{algorithmic}

Deletion, however, requires that we know the position of the clause that we intend to delete. This happens when one of the TAs changes from including to excluding its literal, say literal $l_k$ in clause $C^i_j$. To avoid searching for $C^i_j$ in $L^i_k$, we use $M^{ij}_k$ for direct lookup:  
\vspace{2mm}
\begin{algorithmic}
\centering
\STATE $L^i_k[M^{ij}_k] \gets L^i_k[n^i_k]$
\vspace{2mm}
\STATE $n^i_k \gets n^i_k-1$
\vspace{2mm}
\STATE $M^{i,{L^i_k[M^{ij}_k]}}_k \gets M^{ij}_k$
\vspace{2mm}
\STATE $M^{ij}_k \gets NA$.
\end{algorithmic}
As seen, the clause is deleted by overwriting it with the last clause in the list, and then updating $\mathcal{M}$ accordingly.

\subsubsection{Memory Footprint.} To get a better sense of the impact of indexing on  memory usage, assume that we have a classification problem with $m$ classes and feature vectors of size $o$, to be solved with a TM. Thus, the machine has $m$ classes where each class contains $n$ clauses. The memory usage for the whole machine is about $2 \times m \times n \times o$ bytes, considering a bitwise implementation and TAs with 8-bit memory. The indexing data structure then requires two tables with $m \times o$ rows and $n$ columns. Consequently, assuming 2-byte entries, each of these tables require $2 \times m \times n \times o$ bytes of memory, which is roughly the same as the memory required by the TM itself. As a result, using the indexing data structure roughly triples memory usage.

\subsubsection{Step-by-Step Example.} Consider the indexing configuration in Fig. \ref{fig-index} and the feature vector $\mathbf{x}=(1,0)$. To evaluate the clauses of class 1, i.e., $C_1^+$, $C_1^-$, $C_2^+$, and $C_2^-$ (omitting the class index for simplicity), we initially assume that they all are true. This provides two votes for and two votes against the class, which gives a class score of $0$. We then look up the literals $\lnot x_1$ and $x_2$, which both are false, in the corresponding inclusion lists. We first obtain $C_1^-$ and $C_2^-$ from $\lnot x_1$, which falsifies $C_1^-$ and $C_2^-$. This changes the class score from $0$ to $2$ because two negative votes have been removed from the sum. We then obtain $C_1^-$ and $C_2^-$ from $x_2$, however, these have already been falsified, so are ignored. We have now derived a final class score of $2$ simply by obtaining a total of four clause index values, rather than fully evaluating four clauses over four possible literals.  

Let us now assume that the clause $C_1^+$ needs to be deleted from the inclusion list of $x_1$. The position matrix $\mathcal{M}$ is then used to update the inclusion list in constant time. By looking up $x_1$ and $C_1^+$ for class 1 in $\mathcal{M}$, we obtain the position of $C_1^+$ in the inclusion list $L_1^1$ of $x_1$, which is $1$.  Then the last element in the list, $C_2^+$, is moved to position $1$, and the number of elements in the list is decremented. Finally, the entry $1$ at row $\mathrm{Class~1}:x_1$ and column $C_1^+$ is erased, while the entry $3$ at row $\mathrm{Class~1}:x_1$ and column $C_2^+$ is set to $1$ to reflect the new position of $C_2^+$ in the list. Conversely, if $C_1^+$ needs to be added to e.g. the inclusion list of $x_2$, it is simply added to the end of the $\mathrm{Class~1}:x_2$ list, in position $3$. Then $\mathcal{M}$ is updated by writing $3$ into  row $\mathrm{Class~1}:x_2$ and column $C_1^+$.

\subsubsection{Remarks.} As seen above, the maintenance of the inclusion lists and the position matrix takes constant time. The performance increase, however, is significant. Instead of going through all the clauses, evaluating each and every one of them, considering all of the literals, the TM only goes through the relevant inclusion lists. Thus, the  amount of work  becomes proportional to the size of the clauses. For instance, for the MNIST dataset, where the average length of clauses is about 58, the algorithm goes through the inclusion list for half of the literals (the falsifying ones), each containing about $740$ clauses on average. Thus, instead of evaluating $20~000$ clauses (the total number of clauses), by considering $1568$ literals for each (in the worst case), the algorithm just considers $784$ falsifying literals with an average of $740$ clauses in their inclusion lists. In other words, the amount of work done to evaluate all the clauses with indexing is roughly $0.02$ compared to the case without indexing. Similarly, for the IMDB dataset, the average clause length is about $116$, and each inclusion list contains roughly $232$ clauses on average. Without indexing, evaluating a clause requires assessing $10~000$ literals, while indexing only requires assessing the inclusion lists for $5~000$ falsifying literals. Using indexing reduces the amount of work to roughly $0.006$ of the work performed without indexing. Thus, in both examples, using indexing reduces computation significantly. We now proceed to investigate how the indexing affect inference and learning speed overall.

\section{Empirical Results}\label{sec:empirical_results}
\def\widt{0.75}
\tikzset{
    ultra thin/.style= {line width=0.1pt},
    very thin/.style=  {line width=0.2pt},
    thin/.style=       {line width=0.4pt},
    semithick/.style=  {line width=0.5pt},
    thick/.style=      {line width=0.9pt},
    very thick/.style= {line width=1.4pt},
    ultra thick/.style={line width=2.0pt}
}

To measure the effect of indexing we use three well-known datasets, namely MNIST, Fashion-MNIST, and IMDb. We explore the effect of different number of clauses and features, to determine the effect indexing has on performance under varying conditions.

Our first set of experiments covers the MNIST dataset. MNIST consists of  $28 \times 28$-pixel grayscale images of handwritten digits. We first binarize the images, converting each image into a 784-bit feature vector. We refer to this version of binarized MNIST as M1. Furthermore, to investigate the effect of increasing the number of features on indexing, we repeat the same experiment using 2, 3 and 4 threshold-based grey tone levels. This leads to feature vectors of size 1568 ($= 2 \times 784$), 2352 ($= 3 \times 784$) and 3136 ($= 4 \times 784$), respectively. We refer to these datasets as M2, M3 and M4.

Table~\ref{tab-mnist} summarizes the resulting speedups. Different rows correspond to different number of clauses. The columns refer to different numbers of features, covering training and inference. As seen, increasing the number of clauses increases the effect of indexing. In particular, using more than a few thousands clauses results in about three times faster learning and seven times faster inference.

\begin{table}
\begin{center}
\caption{Indexing speedup for different number of clauses and features on MNIST}
\label{tab-mnist}
\begin{tabular}{|c||c|c||c|c||c|c||c|c|}
\hline
Features & \multicolumn{2}{|c||}{784} & \multicolumn{2}{|c||}{1568} & \multicolumn{2}{|c||}{2352} & \multicolumn{2}{|c|}{3136}\\
\hline
Clauses & Train & Test & Train & Test & Train & Test & Train & Test \\
\hline
1000 & 1.78 & 2.75 & 1.54 & 2.79 & 1.76 & 3.74 & 1.91 & 4.15 \\
2000 & 1.67 & 2.83 &  2.63 & 5.74 & 2.46 & 5.95 & 2.87 & 6.06 \\
5000 & 2.62 & 5.95 & 3.23 & 8.02 & 3.26 & 6.88 & 3.34 & 7.17 \\
10000 & 2.82 & 6.31 & 3.49 & 8.31 & 3.47 & 7.60 & 3.43 & 6.78 \\
20000 & 2.69 & 5.22 & 3.58 & 8.02 & 3.34 & 6.44 & 3.35 & 6.39 \\
\hline
\end{tabular}
\end{center}
\end{table}

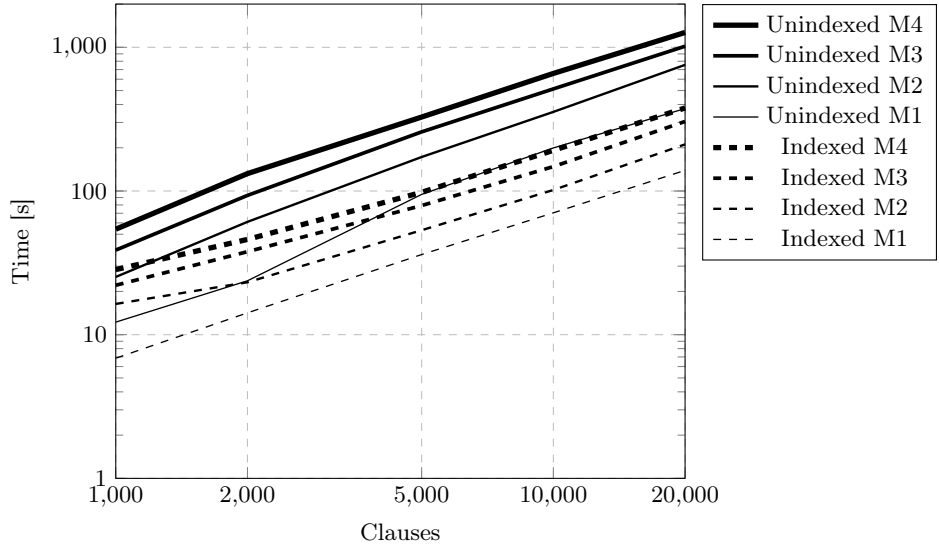
\begin{figure}
    \centering
\begin{tikzpicture}
\begin{axis}[
    xlabel={Clauses},
    ylabel={Time [s]},
    xmin=1000, xmax=20000,
    ymin=1, ymax=2000,
    xtick={1000,2000,5000,10000,20000},
    ytick={1,10,100,1000},
    legend pos=outer north east,
    xmajorgrids=true,
    ymajorgrids=true,
    grid style=dashed,
    xmode=log,
    ymode=log,
    log ticks with fixed point,
	width=\widt * \textwidth
]

\addplot[ultra thick]
    coordinates {(1000, 54.21)(2000, 132.15)(5000, 327.14)(10000, 658.06)(20000, 1268.76)};
\addplot[very thick]
    coordinates {(1000, 38.79)(2000, 93.20)(5000, 257.88)(10000, 513.50)(20000, 1017.59)};
\addplot[thick]
    coordinates {(1000, 25.23)(2000, 60.99)(5000, 172.42)(10000, 355.30)(20000, 755.14)};
\addplot[semithick] 
    coordinates {(1000, 12.23)(2000, 23.72)(5000, 94.69)(10000, 199.41)(20000, 374.86)};

\addplot[ultra thick, dashed]
    coordinates {(1000, 28.38)(2000, 46.03)(5000, 97.92)(10000, 191.85)(20000, 378.46)};
\addplot[very thick, dashed]
    coordinates {(1000, 22.08)(2000, 37.82)(5000, 79.19)(10000, 147.88)(20000, 304.78)};
\addplot[thick, dashed]
    coordinates {(1000, 16.37)(2000, 23.16)(5000, 53.34)(10000, 101.67)(20000, 211.12)};
\addplot[semithick, dashed] 
    coordinates {(1000, 6.86)(2000, 14.19)(5000, 36.12)(10000, 70.49)(20000, 139.21)};
\legend{Unindexed M4, Unindexed M3, Unindexed M2, Unindexed M1, Indexed M4, Indexed M3, Indexed M2, Indexed M1}

\end{axis}
\end{tikzpicture}
\caption{Average training time on MNIST}
\label{fig-mnist-train}
\end{figure}

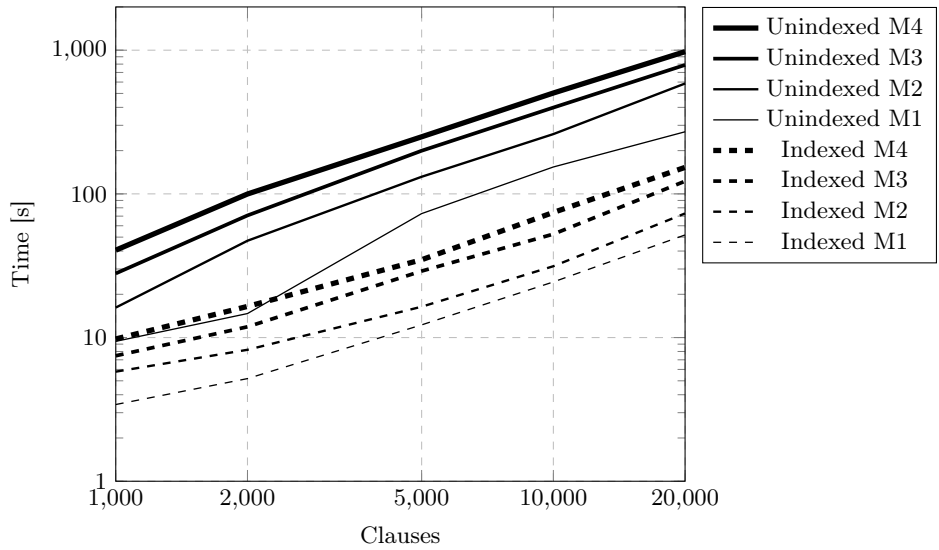
\begin{figure}
    \centering
\begin{tikzpicture}
\begin{axis}[
    xlabel={Clauses},
    ylabel={Time [s]},
    xmin=1000, xmax=20000,
    ymin=1, ymax=2000,
    xtick={1000,2000,5000,10000,20000},
    ytick={1,10,100,1000},
    legend pos=outer north east,
    xmajorgrids=true,
    ymajorgrids=true,
    grid style=dashed,
    xmode=log,
    ymode=log,
    log ticks with fixed point,
	width=\widt * \textwidth
]

\addplot[ultra thick]
    coordinates {(1000, 40.57)(2000, 99.78)(5000, 249.60)(10000, 503.37)(20000, 976.69)};
\addplot[very thick]
    coordinates {(1000, 27.94)(2000, 70.76)(5000, 199.64)(10000, 398.82)(20000, 789.56)};
\addplot[thick]
    coordinates {(1000, 16.17)(2000, 47.18)(5000, 131.53)(10000, 260.64)(20000, 586.37)};
\addplot[semithick] 
    coordinates {(1000, 9.41)(2000, 14.69)(5000, 72.89)(10000, 153.90)(20000, 270.50)};

\addplot[ultra thick, dashed]
    coordinates {(1000, 9.78)(2000, 16.47)(5000, 34.80)(10000, 74.27)(20000, 152.88)};
\addplot[very thick, dashed]
    coordinates {(1000, 7.47)(2000, 11.88)(5000, 29.02)(10000, 52.47)(20000, 122.53)};
\addplot[thick, dashed]
    coordinates {(1000, 5.79)(2000, 8.22)(5000, 16.39)(10000, 31.37)(20000, 73.10)};
\addplot[semithick, dashed] 
    coordinates {(1000, 3.42)(2000, 5.18)(5000, 12.25)(10000, 24.38)(20000, 51.78)};
\legend{Unindexed M4, Unindexed M3, Unindexed M2, Unindexed M1, Indexed M4, Indexed M3, Indexed M2, Indexed M1}

\end{axis}
\end{tikzpicture}
\caption{Average inference time on MNIST}
\label{fig-mnist-test}
\end{figure}

Fig.~\ref{fig-mnist-train} and Fig.~\ref{fig-mnist-test} plots the average time of each epoch across our experiments, as a function of the number of clauses. From the figures, we see that after an initial phase of irregular behavior, both indexed and unindexed versions of the TM suggest a linear growth with roughly the same slopes. Thus, the general effect of using indexing on MNIST is a several-fold speedup, while the computational complexity of the TM algorithm remains unchanged, being linearly related to the number of clauses.

In our second set of experiments, we used the IMDb sentiment analysis dataset, consisting of 50,000 highly polar movie reviews, divided in positive and negative reviews. We varied the problem size by using binary feature vectors of size 5000, 10000, 15000 and 20000, which we respectively refer to as I1, I2, I3 and I4. 

Table~\ref {tab-imdb} summarizes the resulting speedup for IMDb, employing different number of clauses and features. Interestingly, in the training phase, indexing seems to slow down training by about 10\%. On the flip side, when it comes to inference, indexing provided a speedup of up to 15 times.

\begin{table}
\begin{center}
\caption{Indexing speedup for different number of clauses and features on IMDb}
\label{tab-imdb}
\begin{tabular}{|c||c|c||c|c||c|c||c|c|}
\hline

Features & \multicolumn{2}{|c||}{5000} & \multicolumn{2}{|c||}{10000} & \multicolumn{2}{|c||}{15000} & \multicolumn{2}{|c|}{20000}\\
\hline
Clauses & Train & Test & Train & Test & Train & Test & Train & Test \\
\hline
1000 & 0.76 & 1.95 & 0.84 & 1.59 & 0.81 & 1.78 & 0.81 & 2.03 \\
2000 & 1.00 & 4.60 & 0.87 & 3.17 & 0.84 & 4.47 & 0.83 & 4.81 \\
5000 & 0.99 & 8.58 & 0.93 & 8.66 & 0.85 & 9.41 & 0.87 & 9.49 \\
10000 & 1.06 & 15.40 & 0.92 & 11.92 & 0.90 & 13.17 & 0.87 & 13.03 \\
20000 & 1.05 & 15.19 & 0.93 & 15.87 & 0.86 & 13.84 & 0.88 & 13.28 \\
\hline
\end{tabular}
\end{center}
\end{table}

Fig.~\ref{fig-imdb-train} and Fig.~\ref{fig-imdb-test} compare the average time of each epoch across our experiments, as a function of the number of clauses. Again, we observe some irregularities for smaller clause numbers. However, as the number of clauses increases, the computation time of both the indexed and unindexed versions of the TM grows linearly with the number of clauses.

\begin{figure}
    \centering
\begin{tikzpicture}
\begin{axis}[
    xlabel={Clauses},
    ylabel={Time [s]},
    xmin=1000, xmax=20000,
    ymin=10, ymax=10000,
    xtick={1000,2000,5000,10000,20000},
    ytick={10,100,1000,10000},
    legend pos=outer north east,
    xmajorgrids=true,
    ymajorgrids=true,
    grid style=dashed,
    xmode=log,
    ymode=log,
    log ticks with fixed point,
	width=\widt * \textwidth
]

\addplot[ultra thick]
    coordinates {(1000, 416.77)(2000, 834.86)(5000, 2060.76)(10000, 4032.51)(20000, 8114.11)};
\addplot[very thick]
    coordinates {(1000, 277.81)(2000, 554.82)(5000, 1344.77)(10000, 2795.34)(20000, 5377.34)};
\addplot[thick]
    coordinates {(1000, 157.70)(2000, 309.54)(5000, 771.76)(10000, 1537.36)(20000, 3070.37)};
\addplot[semithick] 
    coordinates {(1000, 67.13)(2000, 141.51)(5000, 332.24)(10000, 652.67)(20000, 1286.98)};

\addplot[ultra thick, dashed]
    coordinates {(1000, 513.87)(2000, 1008.55)(5000, 2367.76)(10000, 4629.65)(20000, 9271.15)};
\addplot[very thick, dashed]
    coordinates {(1000, 344.69)(2000, 661.52)(5000, 1584.62)(10000, 3090.07)(20000, 6254.12)};
\addplot[thick, dashed]
    coordinates {(1000, 186.83)(2000, 356.76)(5000, 829.33)(10000, 1665.17)(20000, 3311.09)};
\addplot[semithick, dashed] 
    coordinates {(1000, 88.21)(2000, 141.82)(5000, 336.14)(10000, 616.05)(20000, 1224.135)};
\legend{Unindexed I4, Unindexed I3, Unindexed I2, Unindexed I1, Indexed I4, Indexed I3, Indexed I2, Indexed I1}

\end{axis}
\end{tikzpicture}
\caption{Average training time on IMDb}
\label{fig-imdb-train}
\end{figure}
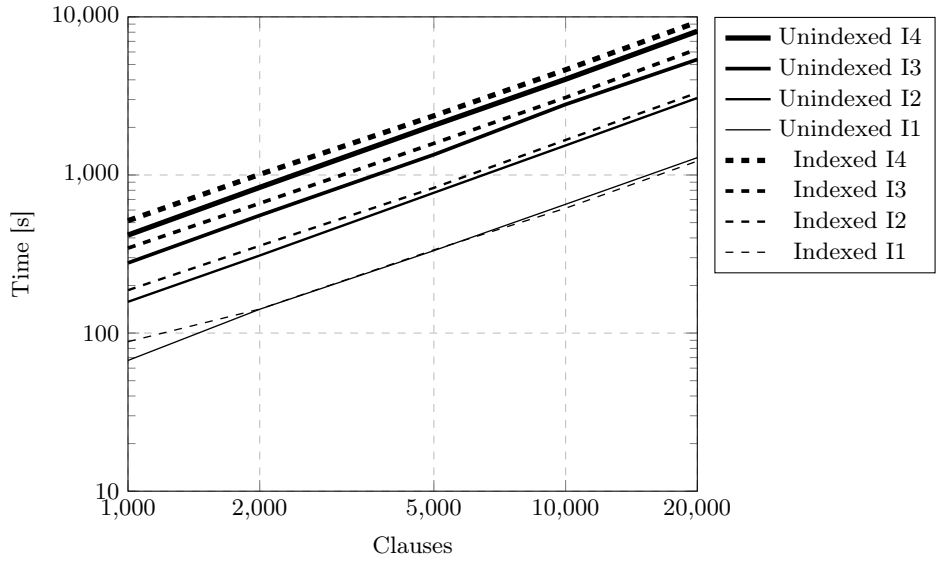

\begin{figure}
    \centering
\begin{tikzpicture}
\begin{axis}[
    xlabel={Clauses},
    ylabel={Time [s]},
    xmin=1000, xmax=20000,
    ymin=1, ymax=2000,
    xtick={1000,2000,5000,10000,20000},
    ytick={1,10,100,1000},
    legend pos=outer north east,
    xmajorgrids=true,
    ymajorgrids=true,
    grid style=dashed,
    xmode=log,
    ymode=log,
    log ticks with fixed point,
	width=\widt * \textwidth
]

\addplot[ultra thick]
    coordinates {(1000, 45.89)(2000, 125.18)(5000, 310.50)(10000, 602.46)(20000, 1130.61)};
\addplot[very thick]
    coordinates {(1000, 30.30)(2000, 87.94)(5000, 240.67)(10000, 487.61)(20000, 944.37)};
\addplot[thick]
    coordinates {(1000, 17.34)(2000, 41.21)(5000, 154.27)(10000, 321.35)(20000, 696.82)};
\addplot[semithick] 
    coordinates {(1000, 10.15)(2000, 27.15)(5000, 92.43)(10000, 210.31)(20000, 409.92)};

\addplot[ultra thick, dashed]
    coordinates {(1000, 22.59)(2000, 26.01)(5000, 32.73)(10000, 46.23)(20000, 85.15)};
\addplot[very thick, dashed]
    coordinates {(1000, 17.02)(2000, 19.68)(5000, 25.58)(10000, 37.02)(20000, 68.23)};
\addplot[thick, dashed]
    coordinates {(1000, 10.88)(2000, 12.99)(5000, 17.81)(10000, 26.96)(20000, 43.90)};
\addplot[semithick, dashed] 
    coordinates {(1000, 5.22)(2000, 5.91)(5000, 10.78)(10000, 13.65)(20000, 26.99)};
\legend{Unindexed I4, Unindexed I3, Unindexed I2, Unindexed I1, Indexed I4, Indexed I3, Indexed I2, Indexed I1}

\end{axis}
\end{tikzpicture}
\caption{Average inference time on IMDb}
\label{fig-imdb-test}
\end{figure}
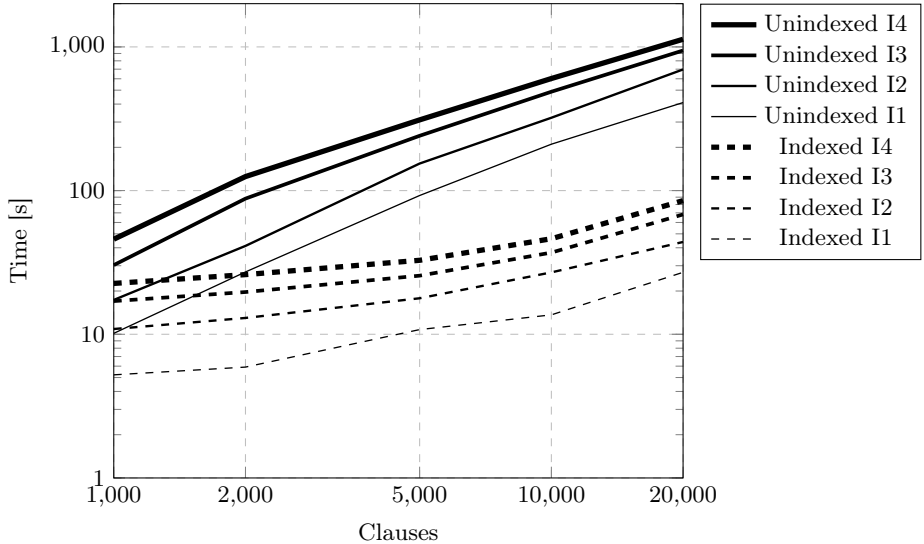

In our third and final set of experiments, we consider the Fashion-MNIST dataset \cite{xiao2017/online}. Here too, each example is a $28 \times 28$-pixel grayscale image, associated with a label from 10 classes of different kinds of clothing items. Similar to MNIST, we binarized and flattened the data using different numbers of bits for each pixel (ranging from one to four bits by means of thresholding). This process produced four different binary datasets with feature vectors of size 784, 1568, 2352 and 3136. We refer to these datasets as F1, F2, F3 and F4, respectively.
The results are summarized in Table~\ref {tab-fashion} for different number of clauses and features. Similar to MNIST, in the training phase, we see a slight speedup due to indexing, except when the number of clauses or the number of features are small. In that case, indexing slows down training to some extent (by about 10\%). On the other hand, for inference, we see a several-fold speedup (up to five-fold) due to indexing, especially with more clauses.

\begin{table}
\begin{center}
\caption{Indexing speedup for different number of clauses and features on Fashion-MNIST}
\label{tab-fashion}
\begin{tabular}{|c||c|c||c|c||c|c||c|c|}
\hline

Features & \multicolumn{2}{|c||}{784} & \multicolumn{2}{|c||}{1568} & \multicolumn{2}{|c||}{2352} & \multicolumn{2}{|c|}{3136}\\
\hline
Clauses & Train & Test & Train & Test & Train & Test & Train & Test \\
\hline
1000 & 0.86 & 0.82 & 0.97 & 1.33 & 1.04 & 1.68 & 1.04 & 1.77 \\
2000 & 0.92 & 1.17 & 1.23 & 2.55 & 1.24 & 2.93 & 1.25 & 3.13 \\
5000 & 1.23 & 2.21 & 1.32 & 2.85 & 1.33 & 2.84 & 1.33 & 2.95 \\
10000 & 1.55 & 3.81 & 1.58 & 4.36 & 1.67 & 4.92 & 1.56 & 3.95 \\
20000 & 1.55 & 3.54 & 1.65 & 4.54 & 1.70 & 4.74 & 1.67 & 5.05 \\
\hline
\end{tabular}
\end{center}
\end{table}

Fig.~\ref{fig-fashion-train} and Fig.~\ref{fig-fashion-test} compare the average time of each epoch as a function of the number of clauses. Again, we observe approximately linearly increasing computation time with respect to the number of clauses.

\begin{figure}
    \centering
\begin{tikzpicture}
\begin{axis}[
    xlabel={Clauses},
    ylabel={Time [s]},
    xmin=1000, xmax=20000,
    ymin=10, ymax=2000,
    xtick={1000,2000,5000,10000,20000},
    ytick={1,10,100,1000},
    legend pos=outer north east,
    xmajorgrids=true,
    ymajorgrids=true,
    grid style=dashed,
    xmode=log,
    ymode=log,
    log ticks with fixed point,
	width=\widt * \textwidth
]

\addplot[ultra thick]
    coordinates {(1000, 88.03)(2000, 188.65)(5000, 467.40)(10000, 795.47)(20000, 1654.04)};
\addplot[very thick]
    coordinates {(1000, 69.50)(2000, 146.46)(5000, 366.88)(10000, 659.71)(20000, 1327.69)};
\addplot[thick]
    coordinates {(1000, 47.39)(2000, 103.65)(5000, 269.09)(10000, 463.13)(20000, 958.90)};
\addplot[semithick] 
    coordinates {(1000, 24.16)(2000, 48.04)(5000, 151.23)(10000, 284.17)(20000, 563.08)};

\addplot[ultra thick, dashed]
    coordinates {(1000, 84.90)(2000, 150.70)(5000, 351.81)(10000, 509.78)(20000, 991.35)};
\addplot[very thick, dashed]
    coordinates {(1000, 67.09)(2000, 117.91)(5000, 276.13)(10000, 394.57)(20000, 779.80)};
\addplot[thick, dashed]
    coordinates {(1000, 48.77)(2000, 84.52)(5000, 203.34)(10000, 292.82)(20000, 582.09)};
\addplot[semithick, dashed] 
    coordinates {(1000, 27.97)(2000, 52.38)(5000, 123.05)(10000, 183.04)(20000, 364.28)};
\legend{Unindexed F4, Unindexed F3, Unindexed F2, Unindexed F1, Indexed F4, Indexed F3, Indexed F2, Indexed F1}

\end{axis}
\end{tikzpicture}
\caption{Average training time on Fashion-MNIST}
\label{fig-fashion-train}
\end{figure}
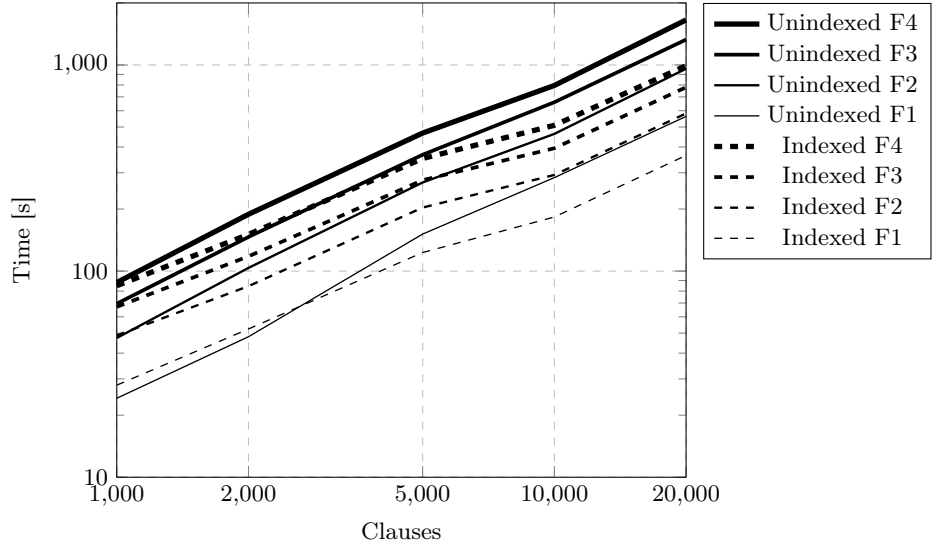

\begin{figure}
    \centering
\begin{tikzpicture}
\begin{axis}[
    xlabel={Clauses},
    ylabel={Time [s]},
    xmin=1000, xmax=20000,
        ymin=1, ymax=1000,
    xtick={1000,2000,5000,10000,20000},
    ytick={1,10,100,1000},
    legend pos=outer north east,
    xmajorgrids=true,
    ymajorgrids=true,
    grid style=dashed,
    xmode=log,
    ymode=log,
    log ticks with fixed point,
	width=\widt * \textwidth
]

\addplot[ultra thick]
    coordinates {(1000, 33.53)(2000, 81.86)(5000, 203.87)(10000, 389.29)(20000, 835.57)};
\addplot[very thick]
    coordinates {(1000, 25.47)(2000, 63.63)(5000, 163.51)(10000, 337.07)(20000, 674.67)};
\addplot[thick]
    coordinates {(1000, 15.33)(2000, 44.72)(5000, 125.34)(10000, 236.01)(20000, 503.26)};
\addplot[semithick] 
    coordinates {(1000, 6.13)(2000, 15.52)(5000, 70.71)(10000, 152.88)(20000, 300.32)};

\addplot[ultra thick, dashed]
    coordinates {(1000, 19.00)(2000, 26.13)(5000, 69.22)(10000, 98.52)(20000, 165.48)};
\addplot[very thick, dashed]
    coordinates {(1000, 15.19)(2000, 21.74)(5000, 57.53)(10000, 68.57)(20000, 142.24)};
\addplot[thick, dashed]
    coordinates {(1000, 11.52)(2000, 17.56)(5000, 43.94)(10000, 54.16)(20000, 110.84)};
\addplot[semithick, dashed] 
    coordinates {(1000, 7.52)(2000, 13.23)(5000, 32.02)(10000, 40.13)(20000, 84.72)};
\legend{Unindexed F4, Unindexed F3, Unindexed F2, Unindexed F1, Indexed F4, Indexed F3, Indexed F2, Indexed F1}

\end{axis}
\end{tikzpicture}
\caption{Average inference time on Fashion-MNIST}
\label{fig-fashion-test}
\end{figure}
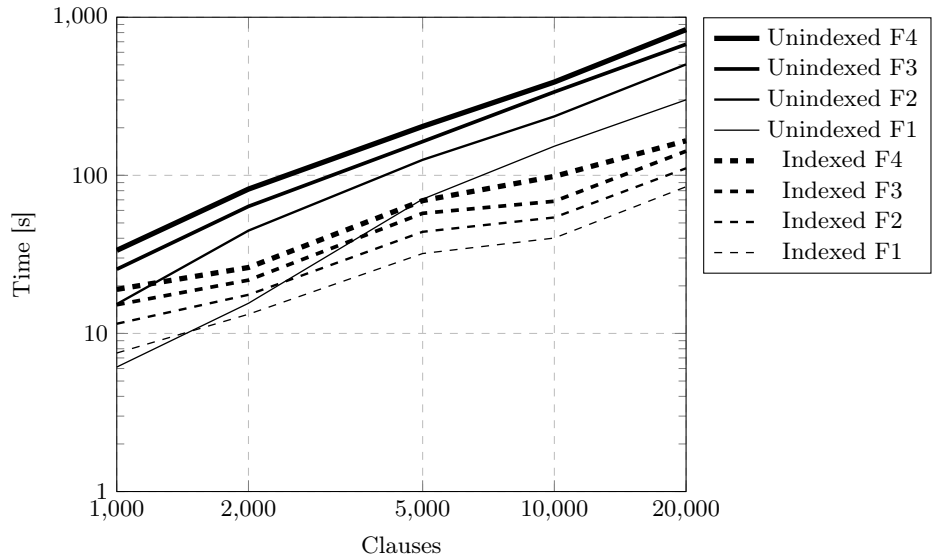

\section{Conclusion and Further Work}\label{sec:conclusion}

In this work, we introduced indexing-based evaluation of Tsetlin Machine clauses, which demonstrated a promising increase in computation speed. Using simple look-up tables and lists, we were able to leverage the typical sparsity of clauses.  We evaluated the indexing scheme on MNIST, IMDb and Fashion-MNIST. For training, on average, the indexed TM produced a three-fold speedup on MNIST, a roughly 50\% speedup on Fashion-MNIST, while it slowed down IMDb training. Thus, the effect on training speed seems to be highly dependent on the data. The overhead from maintaining the indexing structure could counteract any speed gains.

During inference, on the other hand, we saw a more consistent increase in performance. We achieved a three- to eight-fold speedup on MNIST and Fashion-MNIST, and a 13-fold speedup on IMDb, which is reasonable considering there is no index maintenance overhead involved in classification.

In our further work, we intend to investigate how clause indexing can speed up Monte Carlo tree search for board games, by exploiting the incremental changes of the board position from parent to child node.

\bibliographystyle{unsrt}  
\bibliography{references}  
\end{document}